\documentclass[preprint,12pt]{elsarticle}

\usepackage{amssymb}
\usepackage{amsmath}
\usepackage{graphicx}
\usepackage{booktabs}
\usepackage{array}
\usepackage{caption}
\usepackage{multirow}
\usepackage{placeins}
\usepackage[table,xcdraw]{xcolor}

\journal{Aritificial Intelligence in Medicine}

\begin{document}
	
	\begin{frontmatter}
		
		
		
		\title{LGBP-OrgaNet: Learnable Gaussian Band Pass Fusion of CNN and Transformer Features for Robust Organoid Segmentation and Tracking}
		

	\author[label1]{Jing Zhang}
	\ead{zhangjing@uestc.edu.cn}
	
	\author[label1]{Siying Tao}
	\ead{202321050326@std.uestc.edu.cn}
	
	\author[label2]{Jiao Li} 
	\ead{lijiao@alu.scu.edu.cn}
	
	\author[label1]{Tianhe Wang}
	\ead{thwang@std.uestc.edu.cn}
	
	\author[label2]{Junchen Wu}
	\ead{wujc@aimingmed.com}
	
	\author[label1]{Ruqian Hao}
	\ead{ruqian_hao@uestc.edu.cn}
	
	\author[label1]{Xiaohui Du\corref{cor1}}
	\ead{xiaohuie@uestc.edu.cn}
	
	\author[label2]{Ruirong Tan\corref{cor1}}
	\ead{yangmengni@stu.cdutcm.edu.cn}
	
	\author[label3]{Rui Li\corref{cor1}}
	\ead{lirui1@scszlyy.org.cn}
	
	\cortext[cor1]{Corresponding authors.}
	
	 \affiliation[label1]{
		organization={School of Optoelectronic Science and Engineering},
		addressline={University of Electronic Science and Technology of China},
		city={Chengdu},
		postcode={611731},
		state={Sichuan},
		country={China}}
	
	\affiliation[label2]{organization={Translational Chinese Medicine Key Laboratory of Sichuan, Sichuan-Chongqing Joint Key Laboratory of innovation of New Drugs of Traditional, Chinese Medicine Sichuan Institute for Translational Chinese Medicine},
		addressline={Sichuan Academy of Chinese Medicine Sciences},
		city={Chenghu},
		postcode={610041},
		state={Sichuan},
		country={China}}
		
	\affiliation[label3]{organization={Sichuan Cancer Hospital and Institute},
		addressline={Affiliated Cancer Hospital of University of Electronic Science and Technology of China},
		city={Chenghu},
		postcode={610041},
		state={Sichuan},
		country={China}}
		
		\begin{abstract}
			Organoids replicate organ structure and function, playing a crucial role in fields such as tumor treatment and drug screening. Their shape and size can indicate their developmental status, but traditional fluorescence labeling methods risk compromising their structure. Therefore, this paper proposes an  automated, non-destructive approach to organoid segmentation and tracking. We introduced the LGBP-OrgaNet, a deep learning-based system proficient in accurately segmenting, tracking, and quantifying organoids. The model leverages complementary information extracted from CNN and Transformer modules and introduces the innovative feature fusion module, Learnable Gaussian Band Pass Fusion, to merge data from two branches. Additionally, in the decoder, the model proposes a Bidirectional Cross Fusion Block to fuse multi-scale features, and finally completes the decoding through progressive concatenation and upsampling. SROrga demonstrates satisfactory segmentation accuracy and robustness on organoids segmentation datasets, providing a potent tool for organoid research.
		\end{abstract}
		
		\begin{keyword}
			Organoids  \sep Semantic Segmentation \sep Transformer \sep Feature Fusion
			
		\end{keyword}
		
	\end{frontmatter}

\section{Introduction}
\label{intro}
 Organoids are three-dimensional structures that replicate the structure and function of various organs\cite{rossi2018progress}. They encapsulate the cellular heterogeneity, morphological characteristics, and organ-specific functions of diverse tissues, reflecting a composition of cell types that closely resembles physiological conditions. As a vital tool in biological research, organoids are employed in fields such as cancer therapy\cite{joshi2020dna}, drug screening\cite{tong2023rational} and disease modeling\cite{dutta2017disease}.Organoids are a focal point of current and future research in regenerative medicine. The size and shape of organoids can reflect their growth and development, while their internal architecture  indicates cellular organization and regeneration. Therefore, accurate and efficient analysis of organoid-scale images is crucial for evaluating their functionality.

Current advanced optical microscopes enable high spatiotemporal resolution time-lapse observations of organoids. However, traditional manual identification and quantification methods\cite{betjes2021cell} for organoid features are time-consuming and susceptible to subjective bias. Kok\cite{kok2020organoidtracker} proposed a  method combining machine learning and manual correction, but it still depends on human expertise. In 2018, Borten et al.\cite{borten2018automated} developed the OrganoSeg software, which integrates segmentation, filtering, and analysis of 3D bright-field images of cultured organoids. However, the software relies on traditional image processing and threshold segmentation methods which lacks robustness in organoid segmentation with complex backgrounds. The introduction of artificial intelligence offers a reliable solution for automating organoid image analysis\cite{maramraju2024ai}. Nevertheless, the irregular and inconsistent shapes and sizes of organoids, along with their non-smooth edges, complex internal structures, and background noise such as bubbles, pose significant challenges for automated segmentation. . 

Bian, Xuesheng et al.\cite{bian2021deep} proposed an innovative deep neural network (DNN) based on the ResNet backbone, achieving the groundbreaking first automated tracking of organoids by comparing adjacent frames. This approach demonstrates great promise in automating organoid tracking, though it requires high-quality images, which may limit its robustness in more varied conditions. UNet\cite{ronneberger2015u}, widely recognized for its lightweight architecture and exceptional performance in medical image processing, has been further improved in several studies\cite{wang2022novel,matthews2022organoid,hradecka2022segmentation,deininger2023ai} for organoid segmentation and tracking, solidifying its position as a key method in the field. More recently, an organoid segmentation network emphasizing texture information extraction\cite{fan2024local} has been proposed, further enhancing segmentation precision. In addition to CNN-based approaches, TransOrga\cite{qin2023transorga} introduced an innovative organoid segmentation method combining Fourier filtering and the Vision Transformer, showcasing the potential of transformer-based models in this domain. However, despite these advances, the irregular shapes and complex backgrounds in organoid images present ongoing challenges, affecting the accuracy and robustness of these methods.

All of these studies utilize CNN or SwinTransformer-based methods, each excelling in different aspects. CNNs are particularly effective at extracting local features and spatial structural information, while SwinTransformer models capture global context through self-attention mechanisms, offering a complementary strength. Recent research\cite{heidari2023hiformer,yang2023cswin,yuan2023effective} that integrates both SwinTransformer and CNN networks has demonstrated superior performance, showing that combining the strengths of both approaches leads to enhanced results compared to using either model in isolation. Inspired by these promising advancements, we propose the SROrga model for organoid segmentation. This model leverages both CNN and SwinTransformer modules to extract complementary features, which are then fused in the CorrelationBlock proposed in this paper. By integrating these diverse feature extraction strategies, the model effectively handles complex backgrounds with multiple layers of defocusing.Additionally, we integrate multi-scale information from the encoder outputs using the CrossFusion module. Through a self-attention mechanism, larger-scale information is weighted to enhance smaller-scale features, enabling the model to effectively combine the positional information from larger-scale modules with the texture details from smaller-scale modules. Subsequently, a fully connected progressive upsampling structure is employed in the decoder, ensuring comprehensive fusion of information across all levels to generate the final organoid segmentation results. Finally, leveraging the Hungarian algorithm and the target loss compensation algorithm, we successfully accomplished the tracking of organoids over continuous time. In summary, the contributions of this paper are as follows:
\begin{enumerate}
	\item We cultured bladder cancer organoids and curated a dataset for organoid segmentation and tracking. This is the first bladder cancer organoid segmentation dataset with complex backgrounds.
	\item LGBP-Fusion decomposes features into adaptive frequency bands and performs band-wise self-attention, integrating CNN textures with Transformer semantics while reducing cross-modal interference.
	\item We design a Bidirectional Cross Fusion Block (BCF) that leverages bidirectional self-attention and adaptive gating to enable effective information exchange between high-resolution detail.
\end{enumerate}

\section{Dataset}
\label{sec:Dataset}
The data used for training and testing in this study consists of three datasets: our self-constructed mice bladder organoid dataset, public breast organoid dataset, and public brain organoid dataset. By training and testing on these organoid datasets, we can validate the robustness of our model.
\subsection{Self-Constructed Mice Bladder Organoids Dataset}
In this study, the dataset used for analysis consisted of normal bladder epithelium and bladder cancer organoids derived from mice\cite{lees2018tumorevolution}. The normal bladder organoids were sourced from C57/BL6 mice, while the bladder cancer organoids were obtained from BBN-induced muscle-invasive bladder cancer in C57/BL6 mice. Upon dissection, normal bladder and bladder tumor tissues were digested in a solution of 1 mg/mL collagenase at approximately \(37^\circ \)C for 20 minutes to lyse red blood cells. The resulting single-cell suspension was obtained by passing the mixture through a 70$\mu$m pore filter membrane. Subsequently, \(2 \times 10^4\) cells per well were resuspended in a solution consisting of organoid culture medium and Matrigel at a 1:2 ratio. The cell-containing droplets were vertically pipetted into the center of a 24-well plate and incubated at \(37^\circ \)C for 30 minutes until the droplets solidified. Afterward, 600$\mu$L of preheated organoid culture medium was added to each well. The medium was changed every 3 to 4 days, and passaging was performed every 6 to 7 days. During passaging, the organoids were digested with TrypLE enzyme at approximately \(37^\circ \)C for 10 minutes to obtain a single-cell suspension, which was then reseeded in a 24-well plate using Matrigel droplets.The components of the organoid culture medium are listed in Tab\ref{tab:notation}.

The dataset of this study comprises two parts: the segmentation dataset and the tracking dataset. The segmentation dataset consists of bright-field images of bladder and bladder cancer organoids under the same conditions but from different mice and at different passages and cultivation days. These images were captured with a Leica inverted optical microscope at a magnification of 10x (objective). The tracking dataset includes bright-field images taken on days 2 to 4 of cultivation after passaging remove of normal bladder organoids from P8 mice. These images were captured with a Zeiss laser confocal microscope at a magnification of 10x (objective).
\begin{figure}[t]
	\centering
	\includegraphics[width=\textwidth]{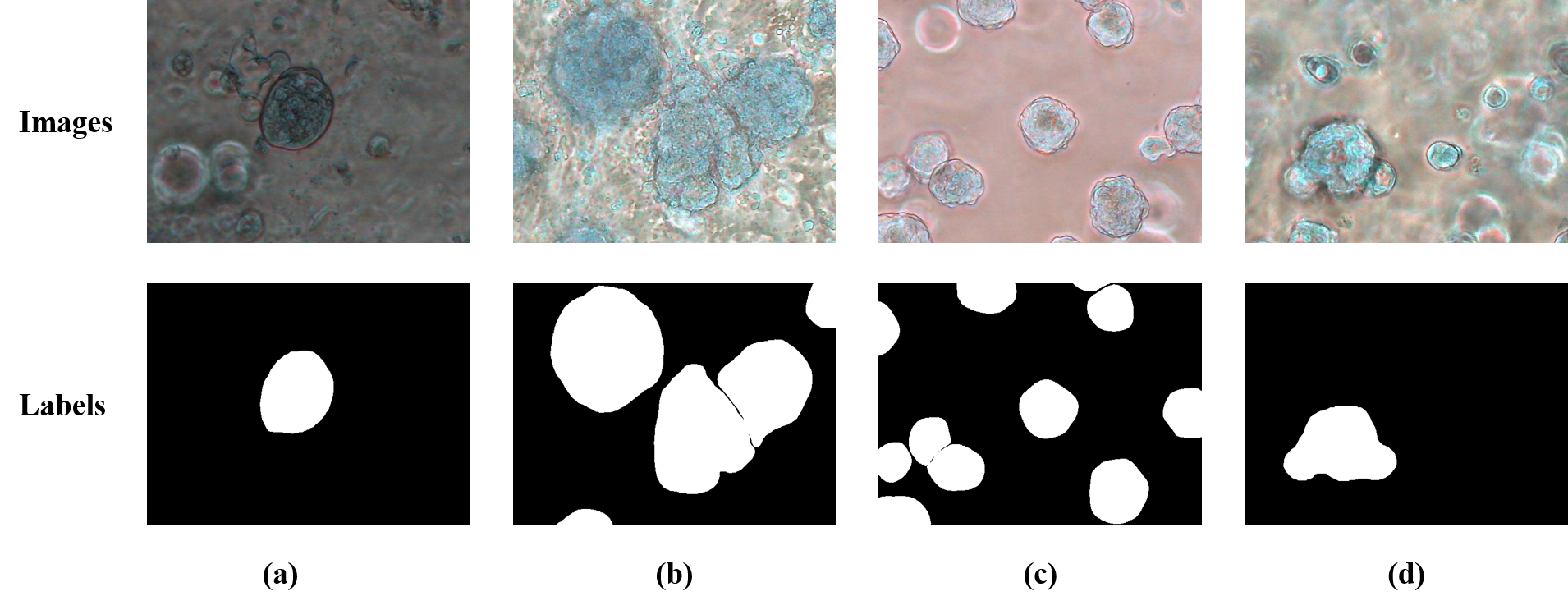}
	\caption{Examples of organoids images and masks}\label{fig:example}
\end{figure}

\begin{figure}
	\centering
	\includegraphics[width=0.4\textwidth]{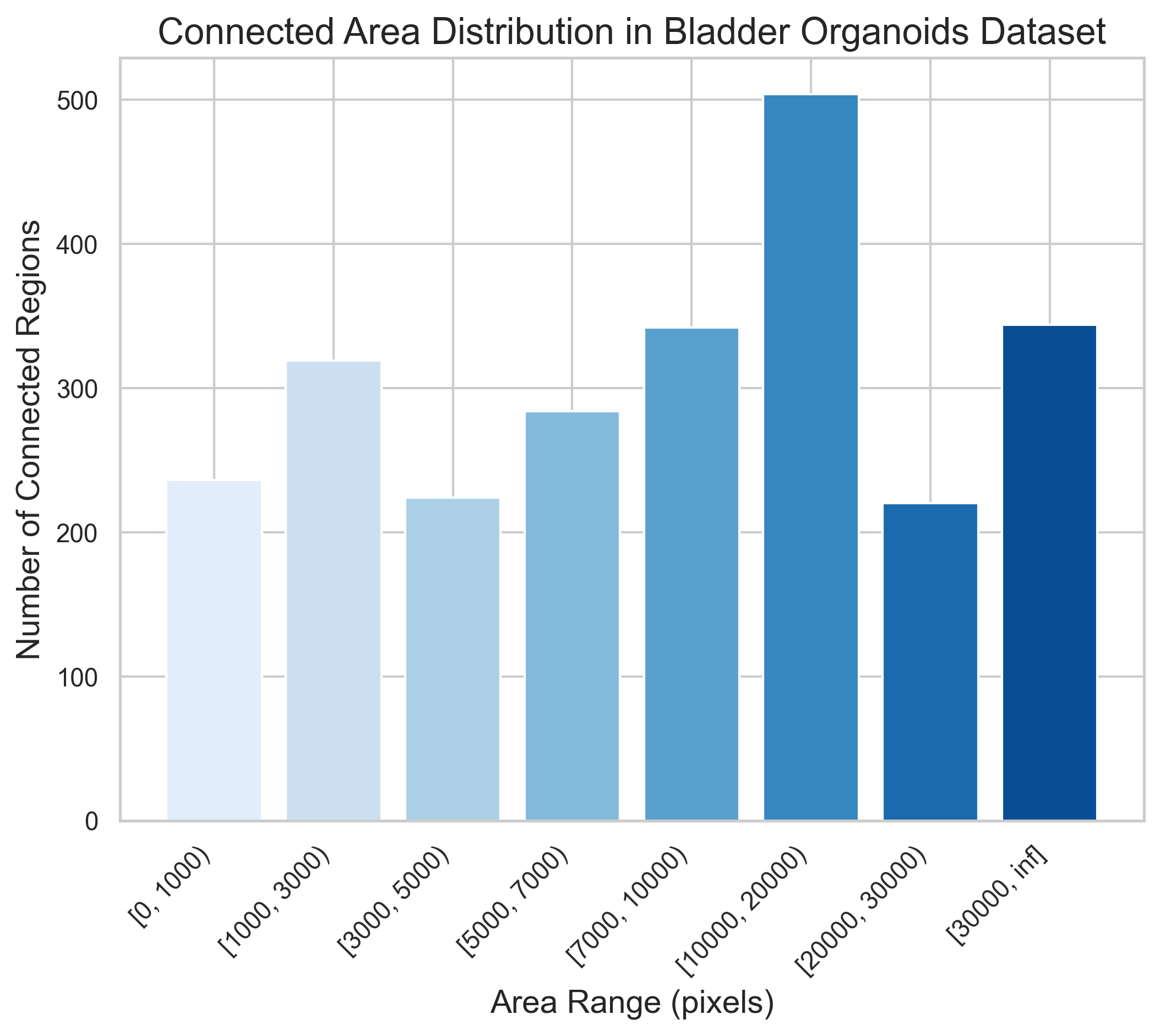}
	\caption{Area Ranges of Mice Bladder Organoids Datasets}\label{fig:datasetarea}
\end{figure}

The segmentation dataset consists of 772 RGB images with a resolution of 648x486 and corresponding masks for organoid segmentation tasks, totaling 2407 organoid regions. The masks are binary images with the same scale as the organoid images and frequency of organoid pixels is 13.91$\%$. The tracking dataset consists of 3 sets of consecutive time images for organoid tracking tasks. Figure \ref{fig:datasetarea} shows the area ranges of this dataset. It indicates that  the number of organoids at each scale in the dataset is evenly distributed. Figure \ref{fig:example} shows some examples of images and corresponding masks in the dataset. This dataset includes organoids of various shapes and sizes, with backgrounds containing numerous bubbles and shadows, posing significant challenges for precise organoid segmentation.
\begin{table*}[t]
	\centering
	\fontsize{8pt}{12pt}\selectfont
	\caption{Medium Composition} \label{tab:notation}
	\begin{tabular}{cc}\hline
		Medium factor & Working concentration   \\\hline
		BM & 72$\%$ \\
		WNT 3A conditioned medium &	20$\%$ \\
		R-Spondin1 conditioned medium &	5$\%$ \\
		Nicotinamide & 10 mM      \\
		A83-01 & 1$\mu$M \\	 
		B27 supplement & 0.5X \\
		N-acetyl-L-cysteine & 1 mM \\
		EGF & 50 ng/mL \\
		Y27632 & 10 $\mu$M \\
		Noggin & 100 ng/mL \\
		FGF7 & 25 ng/mL \\
		FGF10 & 100 ng/mL \\
		\hline
	\end{tabular}
\end{table*}

\subsection{Public Dataset}
\subsubsection{Mammary Epithelial Organoids}
The mammary epithelial organoid dataset was collected by Lucia Hradecká\cite{hradecka2022segmentation} as brightfield two-dimensional time-lapse (2D+t) sequences of mammary epithelial organoid images. It includes two parts: real organoid images and computer-generated bioimage data created using a conditional generative adversarial network (pix2pixHD). We utilize continuous real organoid data for network training and testing. The dataset contains 300 real organoid images, categorized into five types: Cysts, Normally Branching (NB), Massively Branching (MB), Hollowly Branching (HB), and Long Branching (LB). Organoids of different types possess distinct shapes and texture features. Segmenting and tracking these various types of organoids can verify the robustness of the model.
\subsubsection{Brain Organoids}
The brain organoid dataset\cite{schroter2024large} comprises over 1,400 brightfield microscopy images of 64 trackable brain organoids from two different laboratories. The diverse organoid images consist of various clones, frequent imaging time points, and common imaging distractions such as light reflections or shadows caused by the edges of culture plates. We use the brain organoid dataset to validate the robustness of our model, particularly its stability against the impact of imaging distractions.

\FloatBarrier
\section{LGBP-OrgaNet}
\label{sec:LGBP-OrgaNet}

\begin{figure}[t]
	\centering
	\includegraphics[width=\textwidth]{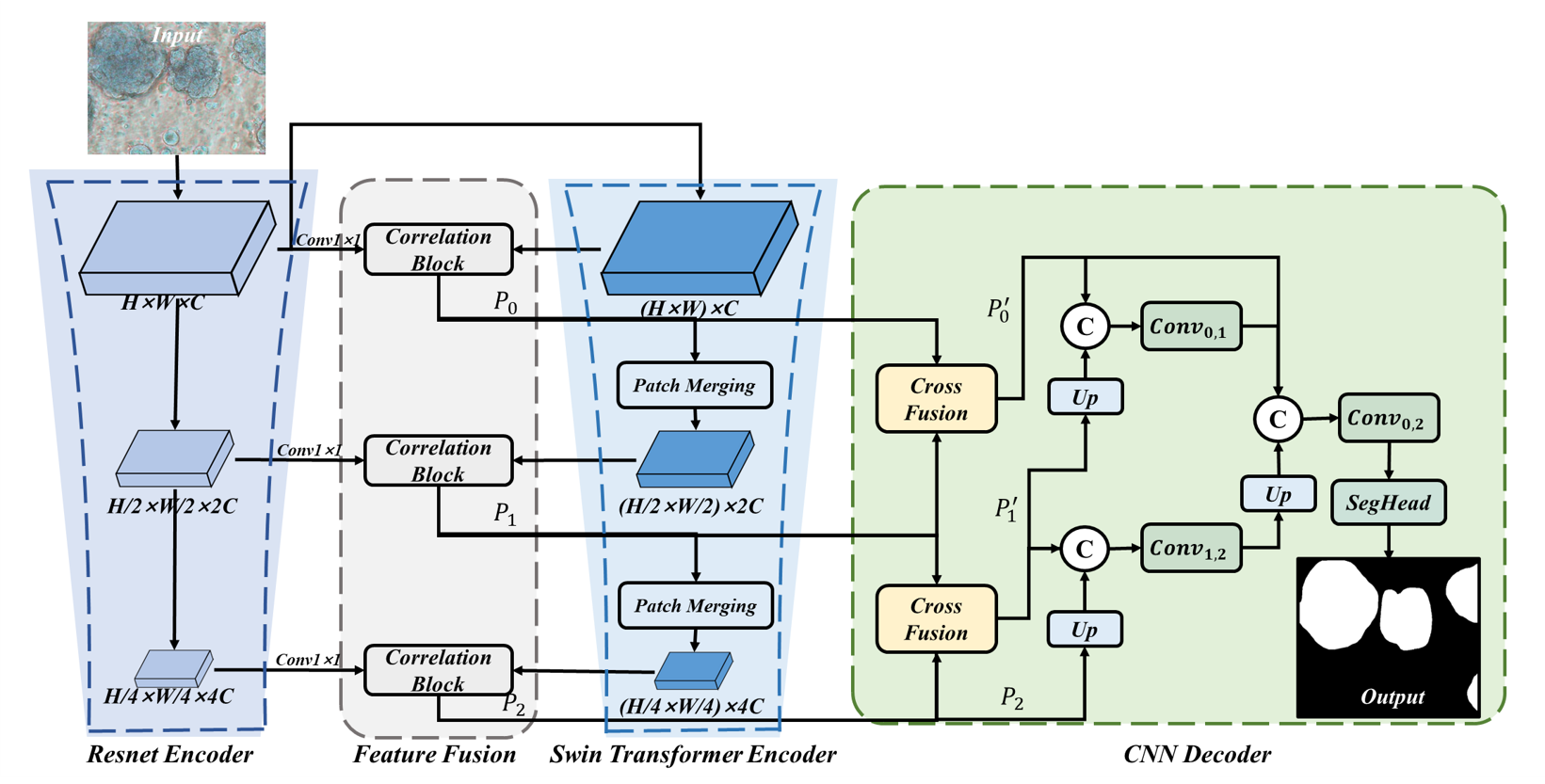}
	\caption{Overview of SROrga structure}\label{fig:overview}
\end{figure}

The overall architecture of the model is illustrated in Figure \ref{fig:overview}, comprising an encoder with a dual-branch structure and a decoder featuring fully connected progressive upsampling. The encoder employs a feature pyramid structure. The ResNet\cite{He2016} branch and SwinTransformer\cite{liu2021swin} branch each contain modules at three different scales. Features at the same scale are integrated through the Correlation module. The LGBP-Fusion module effectively combines features from different domains, outputting the fused features to the subsequent Swin Transformer module and decoder module.
In the decoder, features at adjacent scales, \(P_0\) and\(P_1\), as well as \(P_1\) and \(P_2\) are fused through the Bidirectional Cross Fusion module. This enables the integration of large-scale positional information with fine-grained texture details. Following the fusion, features  \(P_0^\prime\), \(P_1^\prime\) and \(P_2\) are processed through the fully connected progressive upsampling decoder module. The result is passed through the detection head to produce the segmentation output.

\subsection{Dual-encoding Pathway}
\label{secPyramid Features Extraction}
The encoder of the network consists of two branches: the ResNet\cite{He2016} branch and the SwinTransformer\cite{liu2021swin} branch. The ResNet branch focuses on extracting spatial and local features , effectively capturing detailed information and texture features within the images. On the other hand, the Swin Transformer specializes in capturing global information, considering all pixels simultaneously, which helps in handling long-range semantic dependencies. The information from the ResNet and Swin Transformer modules complements each other, contributing to improved accuracy in organoid semantic segmentation.

 The ResNet\cite{He2016} module adopts ResNet50 as the backbone, which addresses the vanishing and exploding gradient problems in the training process of deep neural networks by using residual blocks, making it easier to train deep networks. It consists of an initialization block and three levels of convolutional modules. After each module's output, a 1x1 convolution is applied to align the channels with the transformer module. Images are first encoded by the initialization block before being passed to the SwinTransformer module. This allows the SwinTransformer module to capture global features to extract more information effectively\cite{xiao2021early}.
 
 Swin Transformer\cite{liu2021swin} effectively addresses the computational overhead and limitations in capturing local information encountered by traditional Transformers in computer vision tasks by introducing a local window self-attention mechanism, a shifted window strategy, and a hierarchical architecture. In this paper, the Swin Transformer encoder consists of three modules, with 2, 2, and 6 encoding layers in each module, respectively. The number of detection heads in the encoding layers is 3, 6, and 12, and the window size for the shifted window attention mechanism is set to 7. Between every two Swin Transformer modules, a Patch Merging module is applied. The Patch Merging module merges two adjacent 2x2 patches along the channel dimension into a single patch, performing downsampling of features. 
 
 The outputs of each ResNet module and Swin Transformer module are fused using the CorrelationBlock and then transmitted to the next SwinTransformer module and decoder. The CorrelationBlock fuses the information from both modules through shared weights attention and channel attention, which will be detailed in the following section.

\subsection{Learnable Gaussian Band Pass Fusion}
To address the inherent representational discrepancy between CNN and Transformer features—where CNNs predominantly capture high-frequency local details (e.g., textures and edges), while Transformers emphasize low-frequency global dependencies and semantic structures, we propose the Learnable Gaussian Band Pass Fusion(LGBP-Fusion). Rather than directly performing attention in the spatial domain, the LGBP-Fusion introduces a learnable frequency-band decomposition, which partitions features into multiple frequency components via Gaussian band-pass filters. Cross-modal attention is then independently performed within each frequency band, followed by gated aggregation, enabling fine-grained, controllable, and interpretable feature fusion.
\begin{figure}[t]
	\centering
	\includegraphics[width=\textwidth]{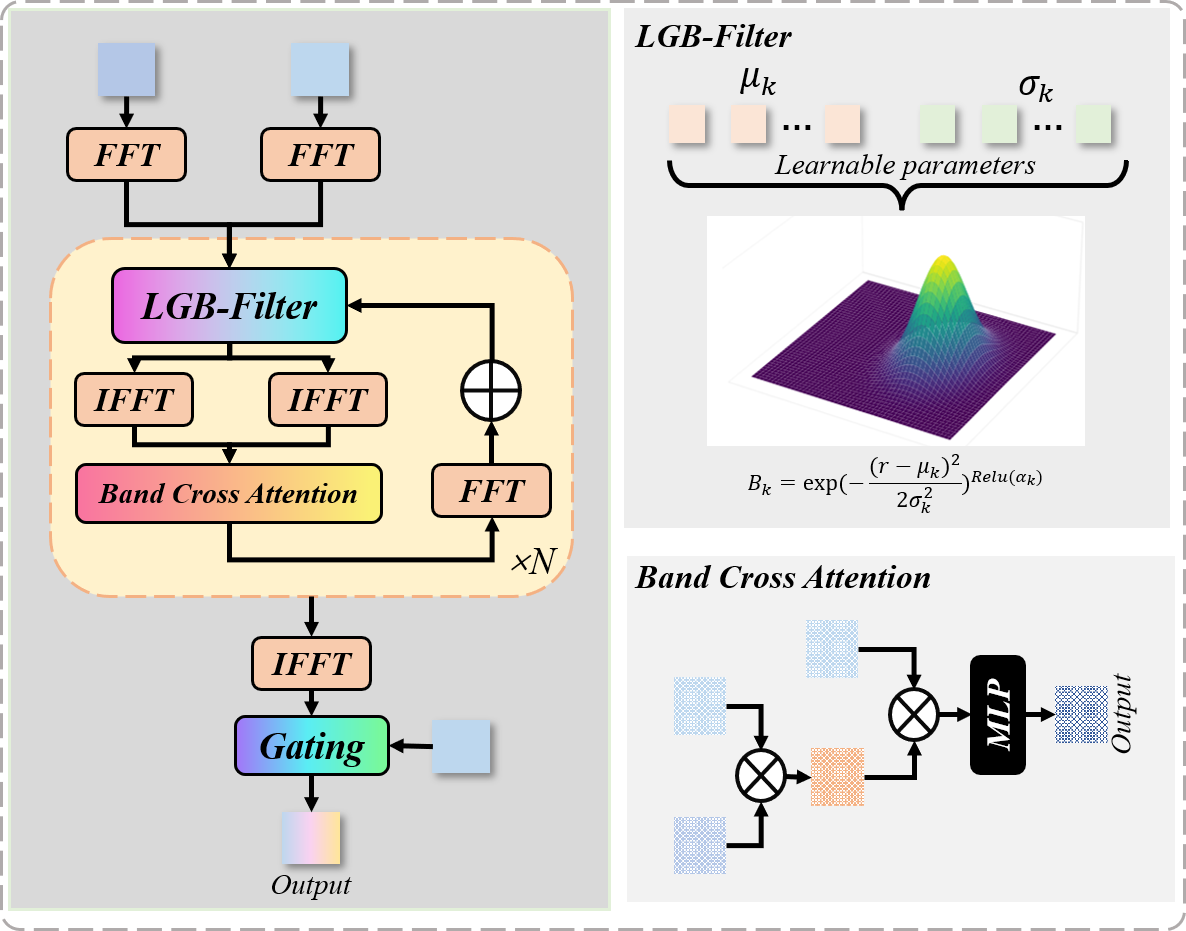}
	\caption{ The overview of Learnable Gaussian Band Pass Fusion(LGBP-Fusion)}\label{fig:LGBP}
\end{figure}
Formally, given CNN features \(F_c\) and Transformer features \(F_t\) , both are first mapped into the frequency domain using a 2D Fourier transform:
\begin{equation}
	\hat{F}_c = \mathcal{F}(F_c), \quad \hat{F}_t = \mathcal{F}(F_t),
\end{equation}

where \(\mathcal{F}(\cdot)\)denotes the Fourier transform. We design a learnable filter bank consisting of \(K\)Gaussian band-pass filters:

\begin{equation}
	B_k(r) = \exp\left(-\frac{(r - \mu_k)^2}{2\sigma_k^2}\right)^{\max(0, \alpha_k)}, \quad k=1,\dots,K,
\end{equation}

where \(r\) is the frequency radius, while \(\mu_k,\sigma_k,\alpha_k\) represent the learnable center frequency, bandwidth, and amplitude factor, respectively. Each filter extracts a sub-band feature as:

\begin{equation}
	\hat{F}_c^{(k)} = \hat{F}_c \odot B_k, \quad \hat{F}_t^{(k)} = \hat{F}_t \odot B_k,
\end{equation}

The filtered features are then transformed back to the spatial domain and fused via BandCrossAttention:
\begin{equation}
	F_c^{(k)} = \mathcal{F}^{-1}(\hat{F}_c^{(k)}), \quad F_t^{(k)} = \mathcal{F}^{-1}(\hat{F}_t^{(k)}),
\end{equation}
\begin{equation}
	Q = W_q F_t^{(k)}, \quad K = W_k F_c^{(k)}, \quad V = W_v F_c^{(k)}
\end{equation}
\begin{equation}
	\text{Attn}(F_t^{(k)},F_c^{(k)}) = \text{softmax}\left(\frac{Q^\top K}{\sqrt{d}}\right)V
\end{equation}
\begin{equation}
	\tilde{F} = \mathcal{F}^{-1}\left(\sum_{k=1}^K \mathcal{F}(\text{Attn}(F_t^{(k)},F_c^{(k)}))\right)
\end{equation}

Finally, we introduce a gating mechanism to adaptively balance the frequency-domain fusion with direct spatial concatenation:
\begin{equation}
	G = \sigma\left(W_g [F_t, \tilde{F}]\right), \quad 
	F_{\text{out}} =G \odot \tilde{F} + (1-G) \odot [F_c,F_t]
\end{equation}
where \(\sigma\) denotes the Sigmoid function, and \(W_g\) is convolutional projections.

This formulation allows CNN and Transformer to complement each other in a frequency-aware manner: low-frequency bands preserve global semantics, high-frequency bands enhance local textures, and mid-frequency bands capture structural details. The learnable nature of \(\mu_k\) and \(\sigma_k\)
enables adaptive adjustment of frequency decomposition, ensuring that the fusion strategy dynamically aligns with task-specific demands while maintaining physical interpretability.

In summary, the LGBP Fusion achieves frequency-aware feature integration by leveraging learnable Gaussian filters to decouple and recombine CNN and Transformer representations across multiple frequency bands. This design not only ensures that global semantics and local details are reinforced in their respective domains, but also provides adaptive flexibility through learnable parameters and clear interpretability through frequency decomposition, thereby yielding more robust and task-aligned feature fusion.
\subsection{Bidirectional Cross Fusion Block}
\begin{figure}[t]
	\centering
	\includegraphics[width=\textwidth]{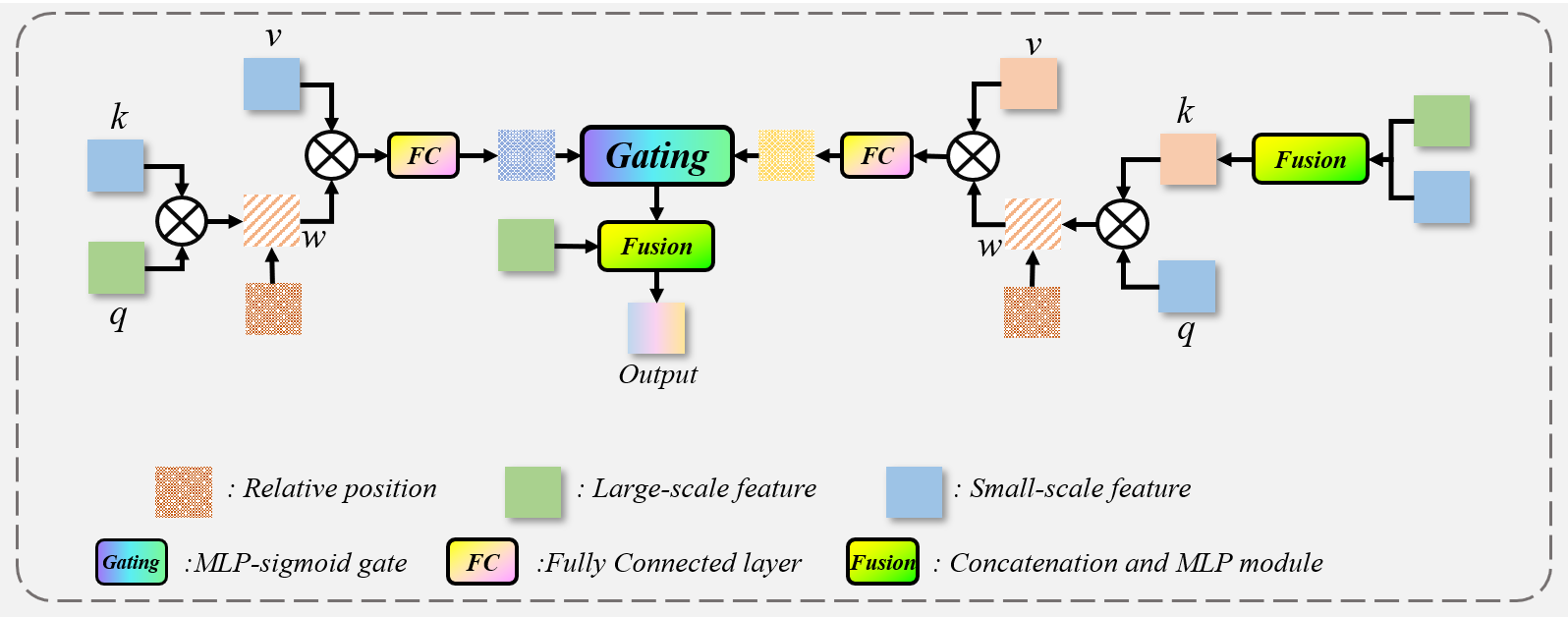}
	\caption{The overview of CrossFusion Block.}\label{fig:BCF}
\end{figure}
To integrate multi-scale features while preserving both spatial details and contextual information, we propose the Bidirectional Cross-Fusion (BCF) module. BCF enables mutual enhancement of high-resolution features $P_1$ and upsampled low-resolution features $P_2$ through bidirectional multi-head attention augmented with learnable relative positional encoding, adaptive gating, and residual connections.

Low-resolution features $P_2$ are first upsampled via a patch expansion interpolator and, together with $P_1$, normalized by LayerNorm. A pre-fusion MLP concatenates these features along the channel dimension, producing cross-scale representations $P_{1,2}$ that enrich attention key-value pairs.

In the high-to-low direction, $P_1$ generates queries $Q_h$, while $P_{2_{up}}$ provides keys $K_h$ and values $V_h$, with attention weights computed as
\begin{equation}
	\text{Attn}{h2l} = \text{softmax} \left( \frac{Q_h K_h^\top}{\sqrt{d}} + B \right),
\end{equation}
where $d$ is the per-head dimension and $B$ is a learnable relative position bias. In the low-to-high direction, $P_{2_{up}}$ generates queries $Q_l$, with $P_{1,2}$ supplying keys $K_l$ and values $V_l$:
\begin{equation}
	\text{Attn}_{l2h} = \text{softmax} \left( \frac{Q_l K_l^\top}{\sqrt{d}} + B \right),
\end{equation}
allowing global context to refine high-resolution features.

The bidirectional outputs $O_{h2l}$ and $O_{l2h}$ are projected and scaled by learnable parameters $\gamma_{h2l}$ and $\gamma_{l2h}$. An MLP-based gate $G \in \mathbb{R}^{B \times N \times 1}$ adaptively fuses them:
\begin{equation}
	F = G \odot O_{h2l} + (1 - G) \odot O_{l2h},
\end{equation}
followed by a residual connection with $P_1$ and a final linear projection on their concatenation, yielding $P_{fused} \in \mathbb{R}^{B \times N \times C}$.

By explicitly modeling bidirectional feature interaction and learnable positional relationships, BCF achieves effective multi-scale fusion with spatial coherence and local-global balance. Its adaptive gating, residual connections, and learnable scaling factors stabilize training, while lightweight parameterization ensures efficiency, establishing BCF as a robust approach for cross-scale feature integration.
\subsection{Decoder}
\label{sec:Decoder}
In the decoder, we employ a fully connected layer-by-layer concatenation upsampling structure. Features $P_0$ and $P_1$ from the encoder, as well as $P_1$ and $P_2$, are respectively fused through the CrossFusion module to obtain enhanced features $P_0^{\prime}$, $P_1^{\prime}$, and $P_2^{\prime}$. After upsampling, $P_1^{\prime}$ is concatenated with $P_0^{\prime}$ and processed by the convolutional module $Conv_{0,1}$ to yield the feature $P_{01}$; $P_2$, once upsampled and concatenated with $P_1^{\prime}$, is then passed through the convolutional module $Conv_{1,2}$ to produce the feature $P_{12}$. Upon further upsampling, $P_{12}$ is concatenated with both $P_{01}$ and $P_0$, and after passing through the final convolutional module and segmentation head, the segmentation results are obtained.

The features in the final layer encompass the shallowest feature $P_0^{\prime}$, the intermediate feature $P_{01}$, and the deepest feature $P_{12}$. By leveraging this multi-layer feature concatenation approach, the model can progressively aggregate image features, thereby enhancing its ability to recognize objects of varying sizes.

%
%

\subsection{Organoid Tracking and Measurement}\label{sec:tracking}

To achieve continuous-time organoid tracking, we first compute the connected regions of the segmentation results and use the Hungarian algorithm to match regions between adjacent frames. Additionally, we employ Kalman filtering to compensate for lost targets.

The Hungarian algorithm is a bipartite graph matching algorithm that finds the optimal match by calculating the matching cost between targets. We compute the cost matrix by using a weighted sum of the region similarity (SSIM) and the Euclidean distance between regions, which can be expressed as Equ\ref{equ:cost},\ref{equ:d}m\ref{equ:s}.

\begin{equation}\label{equ:cost}
	cost(T_i,T_j^{\prime}) = \alpha \cdot d_{euclid}(T_i,T_j^{\prime}) + \beta \cdot s(T_i,T_j^{\prime})
\end{equation}
\begin{equation}\label{equ:d}
	d_{euclid}(T_i,T_j^{\prime}) = \sqrt{(x_i-x_j^{\prime})^2 + (y_i-y_j^{\prime})^2}
\end{equation}
\begin{equation}\label{equ:s}
  s(T_i,T_j^{\prime}) = \frac{\left(2\mu_{T_i}\mu_{T_j'} + C_1\right)\left(2\sigma_{T_i T_j'} + C_2\right)}{\left(\mu_{T_i}^2 + \mu_{T_j'}^2 + C_1\right)\left(\sigma_{T_i}^2 + \sigma_{T_j'}^2 + C_2\right)}
\end{equation}

 There, $T_i,T_j^{\prime}$ represent the $i_{th}$ and $j_{th}$ regions of adjacent frames, respectively, $cost(T_i,T_j^{\prime})$ represents the matching cost between the two regions, $d_{euclid}(T_i,T_j^{\prime})$ is the Euclidean distance between the two regions, and $s(T_i,T_j^{\prime})$ is the similarity between the two regions.
 
 After obtaining the cost matching matrix, we apply the Hungarian algorithm to perform the matching, which provides the matching results between adjacent frames. We set the matching cost between results to be no greater than 20. For unmatched regions, we use Kalman filtering to predict their position in the next frame, preventing the loss of matched targets.
 
 The Kalman filter algorithm is a recursive estimation algorithm used to predict and update the position of a target based on its motion model and observation data. Its basic principle involves two steps for target tracking: prediction and update. In the prediction stage, the Kalman filter uses the target's motion model to predict its current position and velocity; in the update stage, the Kalman filter adjusts the predicted result by minimizing the sum of squared errors, updating the target's state estimate. For unmatched targets, we directly use the Kalman filter's prediction result as a substitute for the actual result to prevent target loss.
 
\begin{figure}[t]
 	\centering
 	\includegraphics[width=\textwidth]{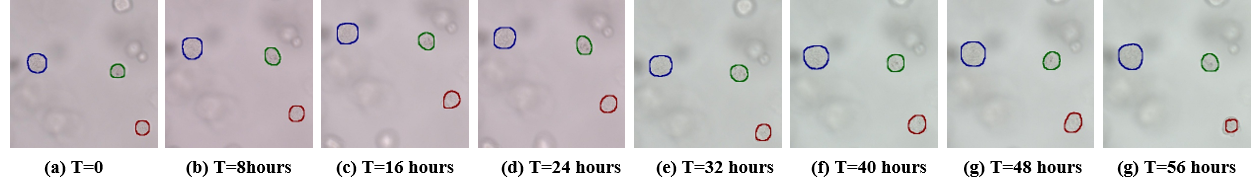}
 	\caption{Organoids tracking results. We use contours of different colors to represent different organoid targets.}\label{fig:trace}
 \end{figure}

\begin{figure}[t]
 	\centering
 	\includegraphics[width=0.6\textwidth]{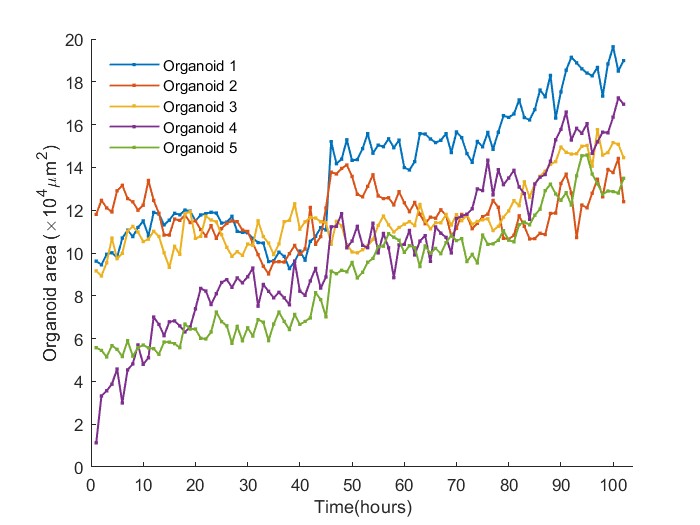}
 	\caption{Organoids area over time.}\label{fig:area}
 \end{figure}

 The tracking results of the organoid images over continuous time are shown in Fig\ref{fig:trace}. It can be observed that our algorithm accurately tracks each organoid target with good robustness. After obtaining the organoid tracking results over continuous time, we can observe the growth of each organoid by calculating their areas. We have recorded the growth of five organoids over a period of 100 hours, as shown in Figure \ref{fig:area}. It can be seen that the areas of the organoids gradually increase while fluctuating. At hour 48, due to the effect of the drug, a step-like growth in area occurs. Observing the dynamic area changes of organoids is of significant importance for studying their growth status and drug effects.
 
 \FloatBarrier

\subsection{Losses}
\label{sec:Losses}
\subsubsection{Isoperimetric Quotient Loss:}
The isoperimetric quotient is defined as the ratio of the square of the perimeter to the area.Isoperimetric Quotient Loss is defined as Equ.\ref{equ:liq}:
\begin{equation}\label{equ:liq}
		L_{IQ} = \frac{{P^2}}{{4\pi A}} \\
		= \frac{{\sum\limits_{i \in \Omega} (dx_i^2 + dy_i^2)}}{{4\pi \sum\limits_{i \in \Omega} {p_i} }}
\end{equation}
Where $P$ represents the perimeter of organoid and $A$ denotes the area of organoid, $dx$ signifies the gradient in the x-direction, $dy$ represents the gradient in the y-direction, and $\omega$ indicates the segmented organoid region. The perimeter is computed from the sum of squared gradients within the perimeter region, while the area is obtained from the sum of pixels within the region. The isoperimetric quotient helps us understand the tightness of the organoid segmentation result. A smaller isoperimetric quotient indicates a more dispersed relationship between the lengths of the perimeter, approaching a circular shape. Through the isoperimetric quotient loss function, the compactness of the image can be measured, leading to results that closely resemble the shape of the organoid, thereby avoiding erroneous segmentation caused by interference from textures in the middle of the organoid.\\
\subsubsection{Dice Loss}
Dice is the intersection of predicted results and true results divided by their union. Dice Loss\cite{milletari2016v} is represented as Equ.\ref{equ:dice}:
\begin{equation}\label{equ:dice}
{L_{Dice}} = 1 - \frac{{2\left| {X \cap Y} \right|}}{{\left| X \right| \cup \left| Y \right|}} = 1 - \frac{{2\left| {\arg \max (X) \cap Y} \right|}}{{\left| {\arg \max \left( X \right)} \right| \cup \left| Y \right|}}
\end{equation}
Where $X$ and $Y$ respectively represent the true and predicted results, $\varepsilon$ is a parameter to prevent instability. $L_{Dice}$ measures the similarity between predicted segmentation results and true segmentation results. It is sensitive to pixel-level prediction results and is suitable for tasks where there is an imbalance between target and background classes in image segmentation. Since organoid data typically have fewer organoid pixels compared to background pixels, using Dice Loss can balance the class proportions.\\
\subsubsection{Focal Loss}
Focal Loss\cite{lin2017focal} is a variant of Cross Entropy Loss with an additional focal parameter. It is represented as Equ.\ref{equ:focal}:
\begin{equation}\label{equ:focal}
L_{Focal} =  - \alpha {(1 - {p_t})^\gamma }\log ({p_t})
\end{equation}
Where $\alpha$ is the factor balancing class weights,$p_t$ is the probability that the model classifies a sample into the corresponding class, and $\gamma$ is the adjustment parameter of the focal factor, which controls the balance between easily classified samples and hard-to-classify samples. When $\gamma=0$, it is equivalent to not introducing focal loss, that is, standard cross-entropy loss; when  $\gamma>0$, the focal loss function will pay more attention to hard-to-classify samples. Focal Loss introduces an adjustment factor to adjust the weights of easily classified samples and hard-to-classify samples, making the loss more focused on hard-to-classify samples. By reducing the contribution of easily classified samples to the loss function, the impact of class imbalance is reduced. This is particularly significant for segmenting organoid pixels and background pixels in a balanced organoid dataset.\\
\subsubsection{Total Loss}
The total loss function can be presented as Equ\ref{equ:total}.. It is obtained by weighting  Isoperimetric Quotient Loss, Dice Loss, and Focal Loss and adding them together. These loss functions enable a balance between the proportions of organoids and background, facilitating easier organoid segmentation while mitigating irregular segmentation caused by interference from textures within the organoids.
\begin{equation}\label{equ:total}
	L = {\lambda _{IQ}}{L_{IQ}} + {\lambda _{Dice}}{L_{D{\rm{ice}}}} + {\lambda _{Focal}}{L_{Focal}}
\end{equation}

\section{Experiment and Evaluation}
\label{sec:Experiment and Evaluation}
In this section, we compare our model with other segmentation models based on both quantitative and qualitative results, and demonstrate the effectiveness of our proposed modules through ablation experiments. The segmentation models used for comparison includes OrganoID\cite{matthews2022organoid}, TransOrga\cite{qin2023transorga}, Unet\cite{ronneberger2015u}, Unet++\cite{zhou2019unet++} and TransUnet\cite{chen2021transunet}.

 OrganoID\cite{matthews2022organoid} is an automatic organoid segmentation and tracking software based on U-Net. TranOrga\cite{qin2023transorga} used a multi-modal feature extraction module to combine spatial and frequency domain features for organoid segmentation. Unet\cite{ronneberger2015u} has achieved good results in medical image segmentation and has also been applied in organoid segmentation\cite{hradecka2022segmentation,deininger2023ai}. Unet++\cite{zhou2019unet++} improves upon the original U-Net by introducing nested skip pathways and deep supervision, which enhances the model's ability to capture fine-grained details. TransUNet\cite{chen2021transunet} combines the strengths of convolutional neural networks and transformers, using a transformer encoder to capture global contextual information, which helps improve segmentation accuracy, especially in complex or large-scale images.  We modified the inputs of these models and trained and tested them on organoids datasets.

\subsection{Details of Experiments}

We implemented our method using PyTorch and trained and tested our model on three organoid segmentation datasets. The input size of our model is set to 224×224, and the batch size is 4. We optimized our model using the stochastic gradient descent (SGD) algorithm with a random weight averaging strategy. The initial learning rate is set to 0.01, and it is reduced by a factor of 0.1 every 10 epochs. We set $\lambda_{Focal}$ to 0.4, $\lambda_{Dice}$ to 0.6, and $\lambda_{IQ}$ to 0.5 and trained our model for 300 epochs on an NVIDIA RTX 3070 GPU with 12 GB of VRAM.

\subsection{Quantitative Results}
We used commonly employed metrics for image segmentation tasks, including accuracy, precision, recall, mean Dice coefficient, IoU, and F1 score, to evaluate segmentation performance. These metrics were illustrated in Table \ref{tab:metric}.

The comparison results between the SROrga model and other semantic segmentation models across three organoid datasets are presented in Table \ref{tab:results1}. As shown, the SROrga model consistently achieves superior or comparable segmentation performance across all datasets.

In the Mice Bladder Organoids dataset, which poses a significant challenge due to large scale variations in the organoids and the complex background, accurately segmenting the organoids is particularly difficult. While other models perform poorly on this dataset, our model achieves a segmentation IoU of 80.91$\%$ and a Dice score of 93.37$\%$, surpassing the second-best model, OrganoID, by 12.54$\%$ and 9.44$\%$, respectively, which demonstrates the robustness of our model.

In the Mammary Epithelial Organoids and Brain Organoids datasets, SROrga still slightly outperforms the others, highlighting its superior segmentation accuracy.

\begin{figure}[t]
 	\centering
 	\includegraphics[width=0.8\textwidth]{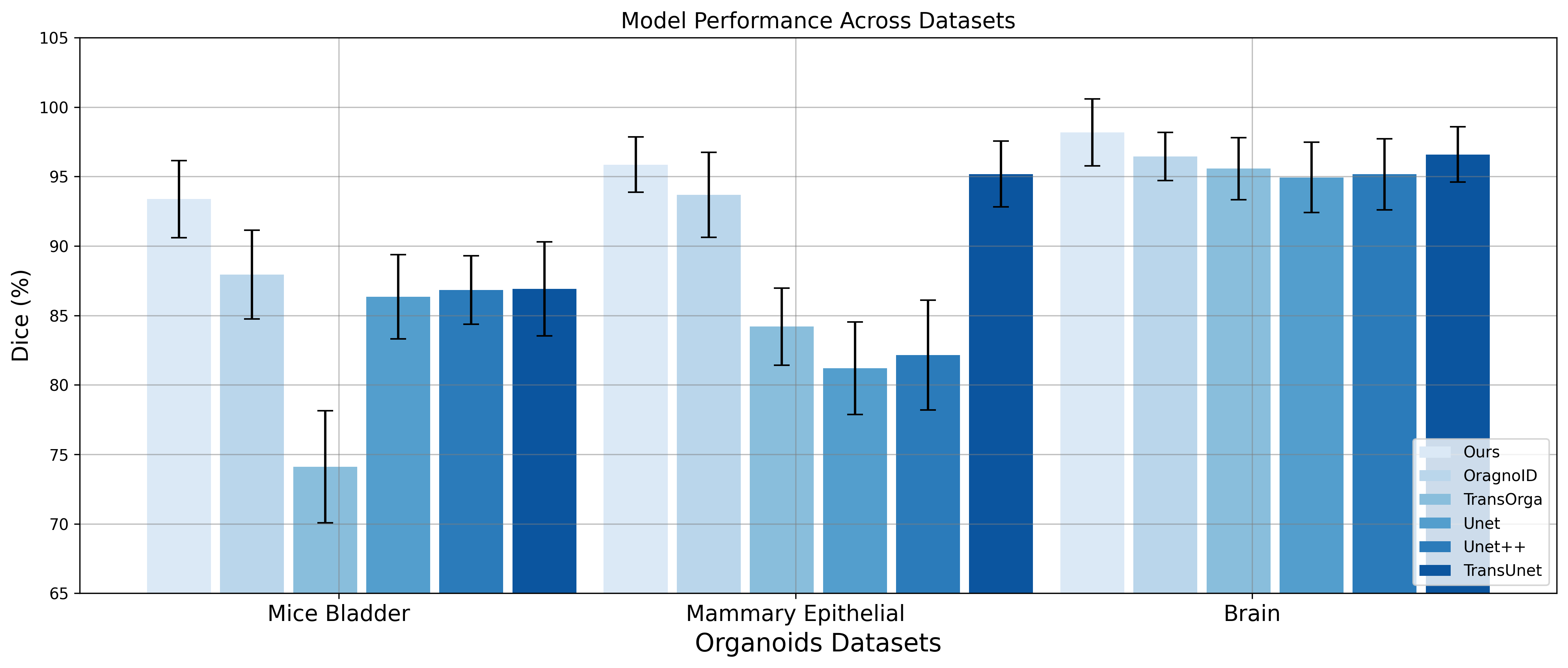}
 	\caption{Segmentation Dice Across Datasets Comparison}\label{fig:dice_std}
 \end{figure}

Figure \ref{fig:dice_std} illustrates the segmentation performance (Dice scores) of various models across three organoid datasets. It is evident that our model significantly outperforms others on the Mice Bladder dataset, demonstrating its superior feature representation capability and robustness when dealing with complex features. On the Mammary Epithelial and Brain datasets, both our model and TransUnet outperform other models, highlighting the critical role of feature fusion between Transformers and CNNs in improving segmentation accuracy. Additionally, our model achieves slightly higher segmentation accuracy than TransUnet, suggesting that the structural design and feature fusion approach of our model provide better expressive power compared to TransUnet.

\begin{figure}[t]
	\centering
	\includegraphics[width=\textwidth]{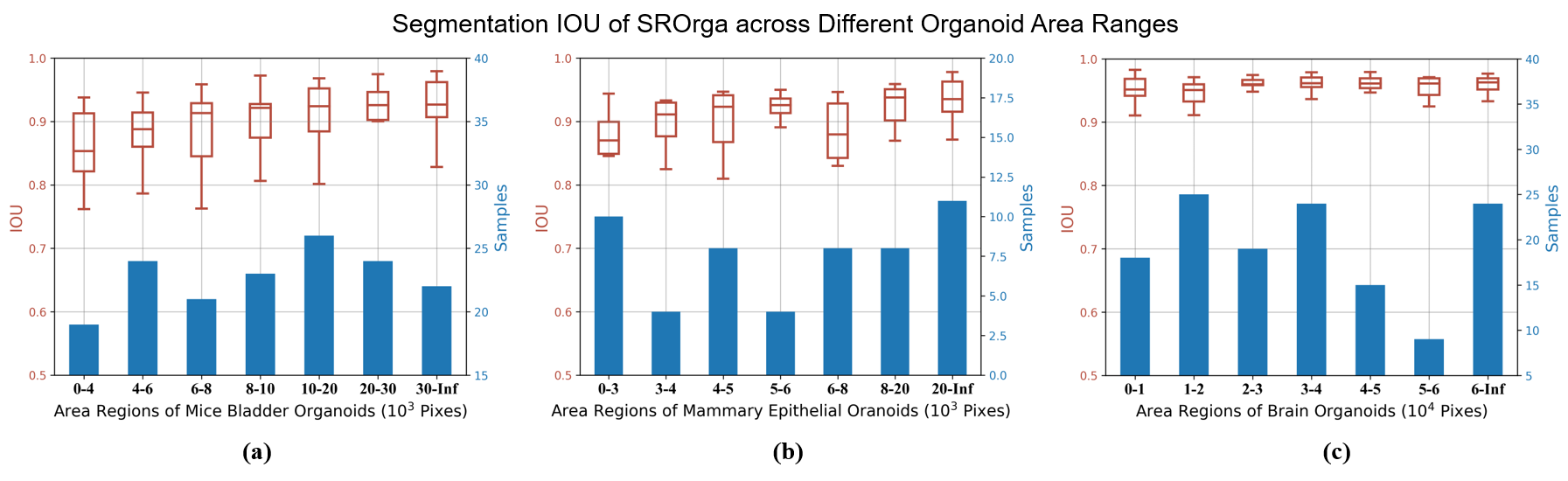}
	\caption{Segmentation IOU of the SROrga across Different Organoid Area Ranges.}\label{fig:iou}
\end{figure}

To further evaluate the robustness of the model in segmenting organoids of varying scales, we analyzed the Intersection over Union (IoU) across organoids at different scales, as shown in Figure \ref{fig:iou}. Specifically, we defined the region of interest as a bounding box extending 10 pixels beyond the actual organoid boundary and calculated the IoU between the segmented and ground-truth regions for each dataset.

In the Mice Bladder Organoids dataset, significant fluctuations in IoU were observed for some samples due to substantial background noise. However, the average IoU across scales remained above 80$\%$, indicating stable performance across features of varying scales. In the Mammary Epithelial Organoids dataset, the model exhibited consistent performance across organoids of different scales, with some minor fluctuations. This suggests that while the segmentation accuracy is satisfactory, there is room for improvement in handling organoids with diverse shapes. In the Brain Organoids dataset, the IoU for all regions exceeded 90$\%$, reflecting the model's strong stability against background interference and lighting variations.


\begin{table*}[t]
	\centering
	\fontsize{8pt}{12pt}\selectfont
	\caption{Metrics} \label{tab:metric}
	\begin{tabular}{cc}\hline
		Metric & Expression\\ \hline
	$TP (True Positive)$ & Correctly identified organoids pixels\\
	$TN (True Negative)$ & Correctly identified background pixels\\
	$FP (False Positive)$ & Incorrectly identified organoids pixels\\
	$FN (False Negative)$ & Incorrectly identified background pixels \\
		$Accuracy$ & $acc = \frac{{TP + TN}}{{TP + TN + EP + FN}}$ \\
		$Precision$ &  $precision = \frac{{TP}}{{TP + FP}}$ \\
		$Recall$ & $recall = \frac{{TP}}{{TP + FN}}$ \\
		$Dice$ & $Dice = \frac{{2 \times \left| {X \cap Y} \right|}}{{\left| X \right| + \left| Y \right|}}$\\
		$Mean Dice$ & $MeanDice = \frac{1}{2}\sum\limits_{i = 0,1} {Dice({X_i})} $\\
		$IOU$ & $IOU = \frac{{TP}}{{TP + FP + FN}}$\\
		$F_1-score$  & ${F_1}-score = \frac{{2 \times precision \times recall}}{{precision + recall}}$ \\ \hline
	\end{tabular}
\end{table*}
 \begin{table}
	\centering
 	\caption{Performance Metrics for Different Segmentation Methods on organoids segmentation datasets} \label{tab:results1}
 	\resizebox{\textwidth}{!}{
 		\begin{tabular}{cccccccc}
 			\hline
 			\textbf{Datasets} & \textbf{Methods} & \textbf{IOU ($\%$)} & \textbf{Precision ($\%$)} & \textbf{Recall ($\%$)} & \textbf{F1-score ($\%$)} & \textbf{Acc ($\%$)} & \textbf{Mean Dice ($\%$)} \\ 
 			\hline
 	\multirow{6}{*}{Mice Bladder Organoids} & SROgra (Ours) & \textcolor{red}{80.91} & \textcolor{red}{90.75} & \textcolor{red}{88.56} & \textcolor{red}{88.53} & \textcolor{red}{97.17} & \textcolor{red}{93.37} \\
 	& OrganoID & \textcolor{blue}{68.31} & \textcolor{blue}{84.42} & 77.44 & \textcolor{blue}{78.78} & \textcolor{blue}{95.38} & \textcolor{blue}{87.93} \\
 	& TransOrga & 42.97 & 56.51 & 63.42 & 54.86 & 88.91 & 74.10 \\
 	& Unet & 64.35 & 75.72 & \textcolor{blue}{83.06} & 76.61 & 93.76 & 86.35 \\
 	& Unet++ & 64.44 & 81.22 & 77.87 & 77.26 & 94.43 & 86.83 \\
 	& TransUnet & 66.81 & 76.73 & 82.24 & 77.09 & 94.83 & 86.90 \\
 	\hline
 	\multirow{6}{*}{Mammary Epithelial Organoids} & SROgra (Ours) & \textcolor{red}{85.89} & \textcolor{blue}{91.24} & \textcolor{red}{93.15} & \textcolor{red}{91.94} & \textcolor{red}{99.57} & \textcolor{red}{95.86} \\
 	& OrganoID & 81.16 & 90.50 & 89.07 & 87.64 & 99.49 & 93.69 \\
 	& TransOrga & 54.90 & 61.27 & 89.12 & 69.30 & 98.23 & 84.19 \\
 	& Unet & 50.33 & 52.89 & 96.29 & 63.71 & 97.44 & 81.19 \\
 	& Unet++ & 52.90 & 59.59 & 73.71 & 63.52 & 98.08 & 81.25 \\
 	& TransUnet & \textcolor{blue}{84.39} & \textcolor{red}{92.40} & \textcolor{blue}{90.17} & \textcolor{blue}{90.59} & \textcolor{blue}{99.55} & \textcolor{blue}{95.18} \\
 	\hline
 	\multirow{6}{*}{Brain Organoids} & SROgra (Ours) & \textcolor{red}{94.49} & \textcolor{red}{97.38} & \textcolor{blue}{96.99} & \textcolor{red}{97.11} & \textcolor{red}{99.08} & \textcolor{red}{98.18} \\
 	& OrganoID & 89.52 & 90.76 & \textcolor{red}{98.49} & 94.24 & 98.20 & 96.45 \\
 	& TransOrga & 86.66 & 90.59 & 95.53 & 92.20 & 98.30 & 95.57 \\
 	& Unet & 85.92 & 90.00 & \textcolor{blue}{95.37} & 91.42 & 97.76 & 94.94 \\
 	& Unet++ & 86.84 & 89.10 & 97.44 & 92.01 & 97.60 & 95.16 \\
 	& TransUnet & \textcolor{blue}{90.06} & \textcolor{blue}{92.65} & \textcolor{blue}{96.99} & \textcolor{blue}{94.35} & \textcolor{blue}{98.41} & \textcolor{blue}{96.59} \\
 	\hline
 			\end{tabular}}
 \end{table}

\subsection{Qualitative Results}
\begin{figure}[t]
	\centering
	\includegraphics[width=\textwidth]{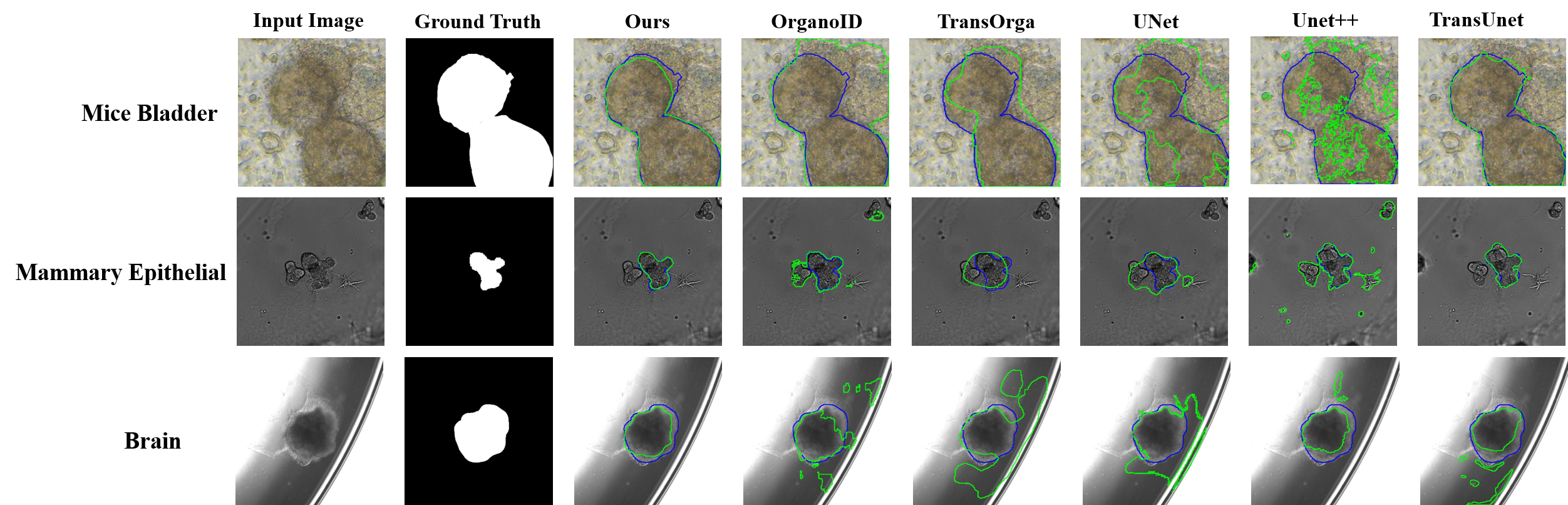}
	\caption{Comparison of segmentation results on organoid segmentation datasets.}\label{fig:segcompare}
\end{figure}

Figure \ref{fig:segcompare} shows segmentation examples of various models on organoid images that are prone to interference. In the Mice Bladder dataset, where organoids overlap and interfere with each other, our model can distinctly separate organoids from interference in the images, whereas other models are affected by background interference, leading to incorrect segmentations. In the Mammary Epithelial dataset, where the subject organoids are disrupted by background impurities, our model effectively separates the interference from the main organoid, accurately capturing the organoid's edges. In contrast, other models fail to recognize the organoid correctly, resulting in erroneous segmentations. In the Brain Organoid, where the organoid contours are very blurred and the contrast with the background is low due to lighting conditions, our model still segments the organoids without interference, outperforming other segmentation models. This demonstrates that our model has strong robustness against interference, achieving accurate organoid segmentation under various challenging conditions.

\FloatBarrier

\subsection{Ablation Experiment}
In this section, we verify the effectiveness of the proposed auxiliary feature extraction ResNet branch, feature fusion module CorrelationBlock, and the incremental upsampling concatenation structure of the decoder. The validation results are shown in Table \ref{tab:ablation}.

\subsubsection{Evaluation of ResNet Branch}
Firstly,to evident the importance of resnet branch, we removed the ResNet branch, allowing information extraction and downsampling only through the SwinTransformer module and PatchMerging module, while keeping the structure of the decoder unchanged, resulting in $Ours\_{ResNet}$. The results show that the IoU of $Ours\_{ResNet}$ decreased by 8.4$\%$, and the Dice score decreased by 3.47$\%$, which show the importance of ResNet Branch. The ResNet module plays a crucial role by assisting the Swin Transformer module in capturing a broader and more diverse set of features, thereby enhancing the model's ability to understand complex patterns.

\subsubsection{Evaluation of LBGP-Fusion}
 Furthermore, to verify the effectiveness of CorrelationBlock, we removed CorrelationBlock and simply added features from SwinTransformer and ResNet before outputting to the next layer, resulting in $Ours\_{Correlation}$. After removing the Correlation Block, the model's segmentation accuracy also decreased, highlighting the importance of the Correlation module in feature fusion. Correlation block allows the model to better integrate information from different sources and improve its representational capacity.
 
 \subsubsection{Evaluation of BCF}
 In addition, to explore the effectiveness of the CrossFusion module, we removed the CrossFusion Block in the experiments and directly input features into the decoder to obtain the result $Ours\_{CorssFusion}$. After removing the CrossFusion module, the model fails to effectively fuse features across different scales, resulting in a decrease in accuracy. CrossFusion Block further refines this process by effectively merging multi-scale features, which directly contributes to a more accurate segmentation output.
 
 \subsubsection{Evaluation of Decoder Structure}
  Finally, to validate the effectiveness of the fully connected decoder structure, with the encoder structure unchanged, we upsampled the small-scale features and added them to the large-scale features, followed by direct output generation through convolutional modules and detection heads, resulting in $Ours\_{Decoder}$. The results shows that the fully connected structure in the decoder better integrates features from different levels, achieving higher accuracy compared to direct convolutional upsampling.

\begin{table}[h]
	\centering
	\caption{Ablation Experients Results.}
	\label{tab:ablation}
	\resizebox{\textwidth}{!}{
		\begin{tabular}{lcccccc}
		\hline
		\textbf{} & \textbf{IOU(\%)} & \textbf{Precision(\%)} & \textbf{Recall(\%)} & \textbf{F1-score(\%)} & \textbf{Acc(\%)} & \textbf{Mean Dice(\%)} \\ \hline
		\textbf{Ours} & 80.91 & 90.75 & 88.56 & 88.53 & 97.17 & 93.37\\
		\textbf{Ours\_ResNet} & 72.51 & 80.79 & 88.30 & 82.82 & 95.32 & 89.90\\
		\textbf{Ours\_LGBP-Fusion} & 75.78 & 87.56  & 85.02 & 84.52 & 96.09 & 91.01 \\
		\textbf{Ours\_BCF} & 65.82 & 79.73 & 78.97 & 77.04 & 94.88 & 86.90 \\
		\textbf{Ours\_Decoder} & 76.18 & 87.35 & 86.16 & 85.19 & 96.06 & 91.34 \\ \hline
		\end{tabular}}

\end{table}
\FloatBarrier

\section{System Application}

The organoid-immune cell co-culture system is a cutting-edge biomedical research technique. By tracking organoids within the co-culture system, we can observe and analyze tumor cell behaviors and responses in the immune microenvironment in real time. This approach allows for a deeper understanding of the interactions between tumors and the immune system and enables the assessment of tumor sensitivity to immunotherapy, which is crucial for cancer treatment.

\begin{figure}[t]
	\centering
	\includegraphics[width=0.6\textwidth]{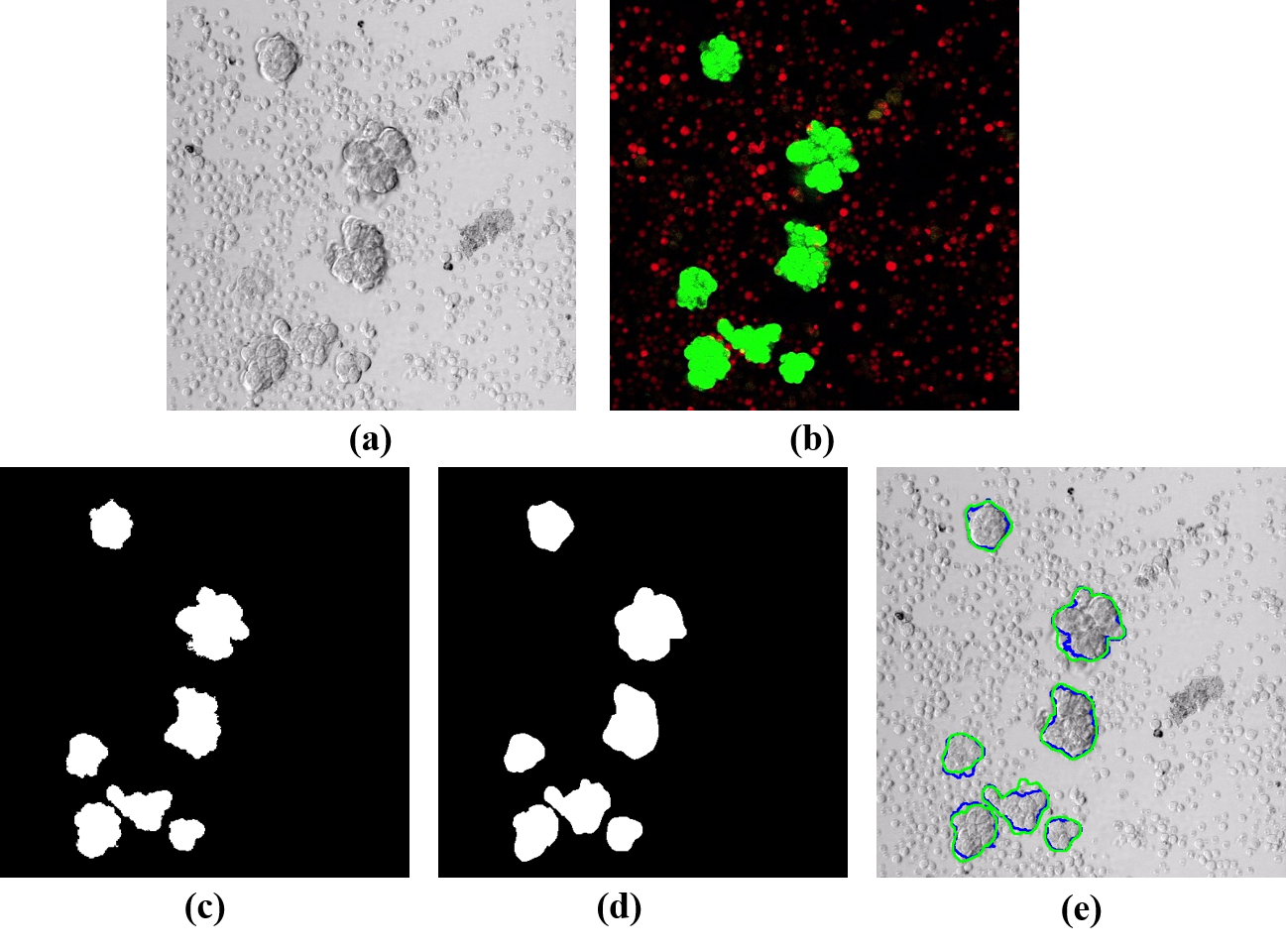}
	\caption{Segmentation of the Immune Cell-Organoid Co-culture System}\label{fig:coculture}
\end{figure}

In this study , we co-cultivate normal bladder organoids or bladder cancer organoids derived from C57/BL6 mice with allogeneic mouse immune cells. Different fluorescent live-cell dyes were used to distinguish between the organoids and immune cells as shown in Fig \ref{fig:coculture}(a)(b). By selecting specific color channels from the fluorescently labeled images, and following by preprocessing and threshold segmentation, we obtained precise contours of the organoids as shown in Fig \ref{fig:coculture}(c). The organoid images under bright-field microscopy and their corresponding segmentation labels constitute the co-cultured organoid dataset.

We utilized a pre-trained model to train and test the co-culture dataset. The results, shown in the Figure \ref{fig:coculture}(d)(e) demonstrate clear and precise segmentation of organoid contours, achieving accurate organoid segmentation and tracking without interference from immune cells, thereby showcasing the robustness of the model. This indicates that we can distinguish between organoids and immune cells without the need for fluorescent labeling, avoiding the interference of fluorescent dyes on organoids, which is of great significance for organoid culture research.
\FloatBarrier
\section{Discussion}

We proposed SROrga, a model that adopts a dual-branch structure with ResNet and SwinTransformer. This hybrid approach enables the model to effectively capture both local spatial features and global contextual information, resulting in superior performance compared to traditional methods. One of the key innovations of our model is the CorrelationBlock, which merges features from both CNN and transformer modules, ensuring a comprehensive feature representation. This fusion mechanism addresses the limitations of individual modules, providing a robust solution for handling the irregular shapes and sizes of organoids as well as background noise. Besides, we also introduces the CrossFusion module for multi-scale feature fusion, which effectively integrates information from different scales through a cross self-attention mechanism. Additionally, the use of multi-scale skip connections in the decoder facilitates effective integration of features at different resolutions, enhancing the accuracy of segmentation.

SROrga demonstrates a significant advancement in the field of organoid segmentation and tracking. Our results show that SROrga outperforms existing models such as OrganoID and TransOrga across various metrics including IOU, precision, recall, F1-score, accuracy, and mean Dice. Even under challenging conditions with complex backgrounds, our model can also accurately segment organoids, indicating that our model has better accuracy and robustness than other models.

In addition, by comparing adjacent frames, we achieved tracking of organoids and observed their growth. We also segmented and tracked organoids in the organoid immune cell co culture system. These indicate that our model plays an important role in the cultivation and research of organoids.

Despite these advancements, there are limitations to our study. Our model still has deficiencies in the tracking algorithm, as it only considers comparisons between adjacent frames without jointly comparing all organoids across the entire time series. This results in reduced robustness of the tracking algorithm. In the future, we plan to propose a multi-frame fusion tracking algorithm to enhance the stability of organoid tracking.

In conclusion, the SROrga model offers a powerful tool for automated organoid segmentation and tracking, with significant implications for biomedical research. By reducing the reliance on manual analysis and fluorescent labeling, it preserves the integrity of organoids and accelerates the research process. Future improvements of the tracking algorithm will further enhance the model's utility and applicability across various research domains.

\section{Conclusion}
\label{sec:Conclusion}
Organoids hold significant importance in biological research. Automated segmentation, tracking, and measurement of organoids can provide insights into their growth, aiding in drug screening and disease treatment research. In this study, we cultured organoids and captured images under a bright-field microscope, which were manually annotated for tracking purposes. Additionally, we captured three sets of sequential images of organoids over time for model tracking. With this dataset, we propose a dual-branch model named SWOrga, which combines Swin Transformer and ResNet. The model introduces Correlation Block to fuse features from ResNet and the Transformer and CrossFusion Block to fuse multi-scale information. In the decoder, the shallowest and deepest features are progressively upsampled and skip-connected for feature fusion and output. By segmenting and comparing regions across sequential organoid images, the model achieves organoid tracking and measurement. SWOrga effectively integrates local texture features from ResNet and long-range semantic features from Swin Transformer, achieving satisfactory segmentation and tracking results on three organoids datasets. This model provides a stable, accurate, and efficient tool for organoid-related medical research. However, our study has limitations; if segmentation errors occur in intermediate frames, the model may lose track of the organoid, indicating a need for improved tracking algorithms. In future research, we will enhance the organoid tracking model to better integrate features from sequential images, improving tracking robustness.

\section{Funding}
This work was supported by the National Natural Science Foundation of China (Grant No. 61405028), and the Fundamental Research Funds for the Central Universities (Grant Nos. ZYGX2019J053 and ZYGX2021YGCX020). The corresponding author for all grants is Jing Zhang.

\section{Author Contributions}
\textbf{Jing Zhang}: Conceptualization, Methodology, Supervision, Writing – original draft, Writing – review \& editing, Funding acquisition.

\textbf{Siying Tao}: Data curation, Formal analysis, Visualization, Investigation, Software, Validation,  Writing – original draft, Writing – review \& editing.

\textbf{Jiao Li}: Data curation, Investigation, Writing – review \& editing.

\textbf{Tianhe Wang}: Methodology,  Validation, Writing – review \& editing.

\textbf{Junchen Wu}: Data curation, Software, Validation, Writing – review \& editing.

\textbf{Ruqian Hao}: Methodology, Validation, Writing – review \& editing.

\textbf{Xiaohui Du}: Supervision, Project administration, Writing – review \& editing.

\textbf{Ruirong Tan}: Data curation, Resources, Investigation, Writing – review \& editing.

\textbf{Rui Li}: Data curation, Resources, Writing – review \& editing.\\

\section{Declaration of Competing Interest}
The authors declare that they have no known competing financial interests or personal relationships that could have appeared to influence the work reported in this paper.


%
%


\bibliographystyle{unsrt}
\bibliography{reference}
\end{document}